\documentclass[conference]{IEEEtran}
\IEEEoverridecommandlockouts

\usepackage[english]{babel}
\usepackage{amsmath}
\usepackage{graphicx}
\usepackage{textcomp}
\usepackage{xcolor}
\usepackage{tikz} 
\usepackage{pgfplots} 
\usepackage{float}
\usepackage{longtable}
\usepackage{multirow}
\usepackage{algorithm}
\usepackage{algpseudocode}
\usepackage{booktabs}
\usepackage{todonotes}
\usepackage{balance}
\usepackage{soul}		
\usepackage{dblfloatfix}
\usepackage{subcaption}
\usepackage{nameref}
\usepackage{todonotes}
\usepackage{hyperref}

\def\BibTeX{{\rm B\kern-.05em{\sc i\kern-.025em b}\kern-.08em
    T\kern-.1667em\lower.7ex\hbox{E}\kern-.125emX}}
    
\graphicspath{{./img/}}

\usepgfplotslibrary{units}

\pgfplotsset{compat = 1.3}

\algblock{ParFor}{EndParFor}
\algnewcommand\algorithmicparfor{\textbf{parfor}}
\algnewcommand\algorithmicpardo{\textbf{do}}
\algnewcommand\algorithmicendparfor{\textbf{end\ parfor}}
\algrenewtext{ParFor}[1]{\algorithmicparfor\ #1\ \algorithmicpardo}
\algrenewtext{EndParFor}{\algorithmicendparfor}

\algrenewcommand\alglinenumber[1]{\tiny #1:}

\usepackage{environ}
\NewEnviron{scaled_IEEEeqnarray}{%
	\begin{IEEEeqnarray}{l}
		\scalebox{0.8}{$\BODY$}
	\end{IEEEeqnarray}
}

\begin{document}

\title{Integration of a systolic array based hardware accelerator into a DNN operator auto-tuning framework \\ \thanks{The research leading to these results was supported by the Competence Center Karlsruhe for AI Systems Engineering (CC-KING \url{https://www.ai-engineering.eu}) sponsored by the Ministry of Economic Affairs, Labour and Tourism Baden-Württemberg. It was conducted within the project KIsSME (Artificial Intelligence for selective near-real-time recordings of scenario and manoeuvre data in testing highly automated vehicles, Grant No. 19A20026C), funded by the German Federal Ministry for Economic Affairs and Climate Action.}}

	
\author{
	\IEEEauthorblockN{Federico Nicolás Peccia}
	\IEEEauthorblockA{FZI Research Center for Information Technology\\
		Karlsruhe, Germany \\
		peccia@fzi.de}
	\and
	\IEEEauthorblockN{Oliver Bringmann}
	\IEEEauthorblockA{University of Tübingen\\
		Tübingen, Germany \\
		oliver.bringman@uni-tuebingen.de}
	}

		

	
\maketitle


	
\begin{abstract}
	 The deployment of neural networks on heterogeneous SoCs coupled with custom accelerators is a challenging task because of the lack of end-to-end software tools provided for these systems. Moreover, the already available low-level schedules and mapping strategies provided by the accelerator developers for typical tensor operations are not necessarily the best possible ones for each particular use case. This is why frameworks which automatically test the performance of the generated code on a specific hardware configuration are of special interest. In this work, the integration between the code generation framework TVM and the systolic array-based accelerator Gemmini is presented. A generic schedule to offload the GEneral Matrix Multiply (GEMM) tensor operation onto Gemmini is detailed, and its suitability is tested by executing the AutoTVM tuning process on it. Our generated code achieves a peak throughput of 46 \textit{giga-operations per second} (GOPs) under a 100 MHz clock on a Xilinx ZCU102 FPGA, outperforming previous work. Furthermore, the code generated by this integration was able to surpass the default hand-tuned schedules provided by the Gemmini developers in real-world workloads.
\end{abstract}

\begin{IEEEkeywords}
	RISC-V, FPGA, TVM, Gemmini, Accelerator, Code Generation
\end{IEEEkeywords}

\section{Introduction}
\label{section:introduction} 

Heterogeneous SoCs generators like Chipyard \cite{chipyard} or HEROv2 \cite{DBLP:journals/corr/abs-2201-03861} are quickly being adopted for an increasing amount of use cases thanks to their great adaptability. This kind of generator exposes an enormous amount of possible configurations to the user, thus enabling the generation of SoCs tailor-made for specific applications. Thanks to the open nature of these projects, hardware accelerators like Gemmini \cite{gemmini-dac}, VTA \cite{moreau2018} or NVDLA \cite{nvdla} are being developed to offload specific workloads from the CPU, allowing the deployment of complex algorithms onto edge platforms.

Although the developers of these accelerators normally provide hand-tuned kernels to offload commonly used tensor operations into the accelerator, the high configurability of these SoC generators becomes a problem: there is no guarantee that these default schedules will provide the best throughput across all possible SoC configurations.

This is why automatic code generation and evaluation tools are becoming increasingly popular \cite{ansor,tvm,flexbuffer}. These present enormous advantages, as they can easily generate different mapping options for each tensor operator and test them to automate the process of finding the best schedule parameters given a particular SoC configuration (a process known as \textit{auto-tuning}). By measuring on a physical hardware platform, the impact of other system components on the accelerator's operation is also taken into account. 

To demonstrate the advantages of this auto-tuning process for hardware accelerators, this work presents the integration of the Gemmini accelerator into the TVM deployment framework. The paper is organized as follows: first, a scheduling search space for a GEMM operation on a generic systolic array accelerator is proposed in Section \ref{section:scheduling}. Then, Section \ref{section:tvm} presents the integration of the Gemmini accelerator into the TVM framework. \footnote{As of the date of submission of this paper, the merging of this integration into the main TVM branch is still in progress}. Finally, Section \ref{section:experiments} compares the auto-tuning of different GEMM workloads against a reference implementation \cite{xu2020}, and demonstrates that our scheduling definition achieves improvements in terms of \textit{giga-operations per second} (GOPs) for all workloads. The measurements are expanded by presenting the autotuning results for operators extracted from the Baidu DeepBench Workloads, outperforming the default handcrafted kernels provided by the Gemmini developers.

\begin{figure*}[!ht]
	\scalebox{0.9}{
		\centering
		\begin{subfigure}[b]{0.18\textwidth}
			\centering
			\includegraphics[width=0.8\textwidth]{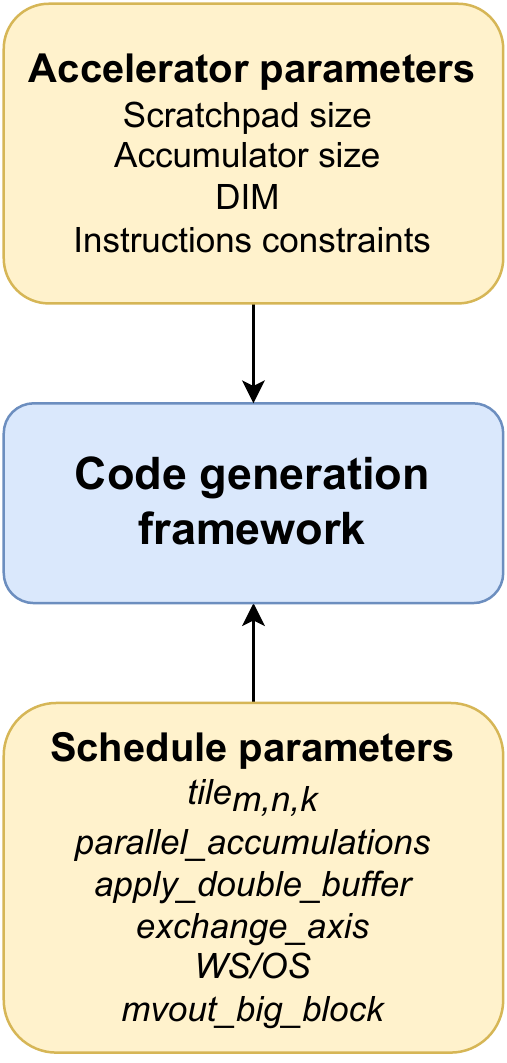}
			\caption{}
			\label{fig:params}
		\end{subfigure}
		\quad
		\begin{subfigure}[b]{0.285\textwidth}
			\centering
			\begin{algorithm}[H]
				\tiny
				\begin{algorithmic}[1]
					\State config.accel()
					\For{$i_o=0$ to $(M/tile_{m_{1}})-1$}
					\For{$j_o=0$ to $(N/tile_{n_{1}})-1$}
					\State move.in($D'$)
					\For{$k_o=0$ to $(K/tile_{k_{1}})-1$}
					\State move.in($A'$)
					\State move.in($B'$)
					\For{$i_i=0$ to $(tile_{m_{1}}/tile_{m_{2}})-1$}
					\For{$j_i=0$ to $(tile_{n_{1}}/tile_{n_{2}})-1$}
					\For{$k_i=0$ to $(tile_{k_{1}}/tile_{k_{2}})-1$}
					\State gemm.DIMxDIM($C'$,$A'$,$B'$)
					\EndFor
					\EndFor
					\EndFor
					\EndFor
					\For{$i_i=0$ to $(tile_{m_{1}}/tile_{m_{2}})-1$}
					\For{$j_i=0$ to $(tile_{n_{1}}/tile_{n_{2}})-1$}
					\State move.out($C'$)
					\EndFor
					\EndFor
					\EndFor
					\EndFor
				\end{algorithmic}
			\end{algorithm}
			\caption{}
			\label{fig:pseudocode}
		\end{subfigure}
		\begin{subfigure}[b]{0.56\textwidth}
			\includegraphics[width=\textwidth]{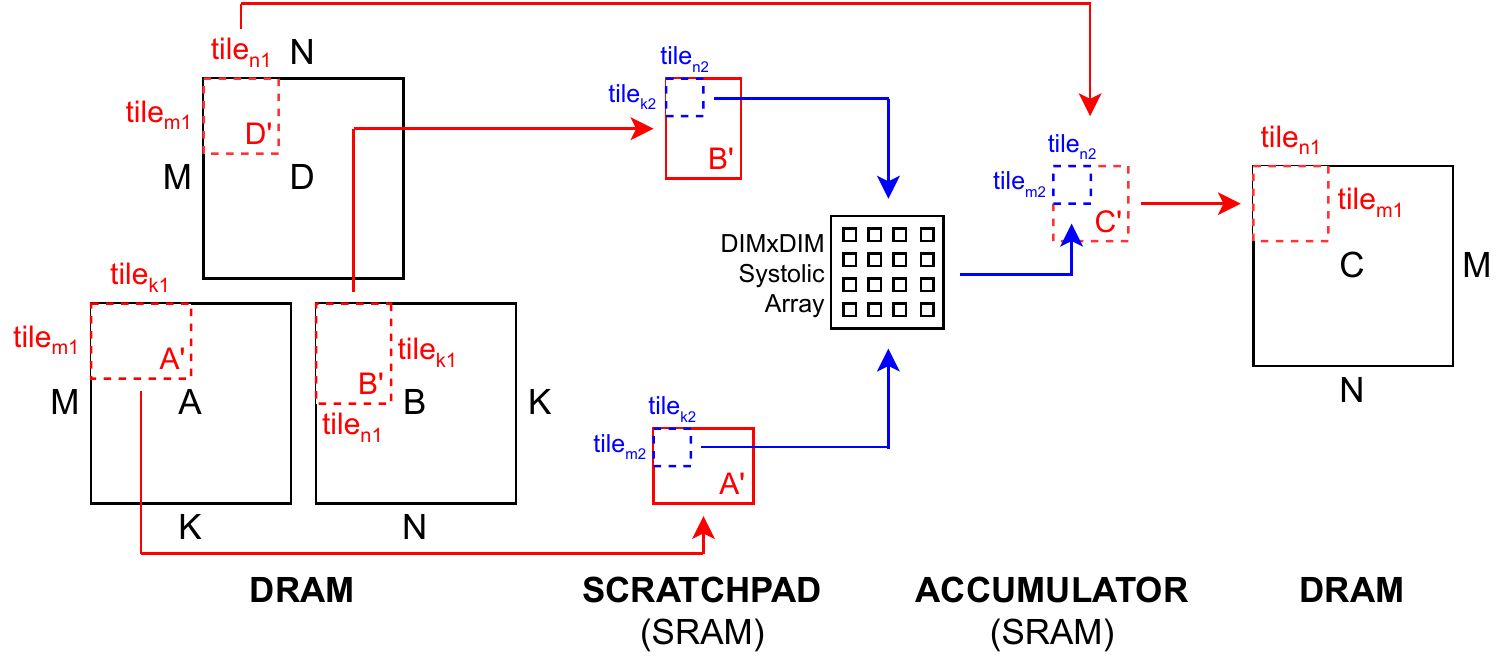}
			\caption{}
			\label{fig:graph_repr}
		\end{subfigure}
	}
	\caption{(\protect{\subref{fig:params}}) shows the proposed GEMM schedule parameters for an accelerator based on a $DIM\times DIM$ systolic array able to execute a generic GEMM with form $C = A\times B + D$. (\protect{\subref{fig:pseudocode}}) shows an example generated pseudocode for the operation, and (\protect{\subref{fig:graph_repr}}) shows a graphical representation of the move of data in and out of the accelerator.}
	\label{fig:gemm_schedule}
\end{figure*}

\section{Scheduling a GEMM operation on a systolic array}
\label{section:scheduling}

In 1982, Kung \cite{kung1982} presented the advantages of systolic array architectures for the execution of matrix operations. These have been especially attractive to accelerate neural network operators because of their high data reuse. Several accelerators were built using systolic arrays (or similar processing element distributions) as its core \cite{chen2017,chen2019,liu2020,wei2017,xu2021,jouppi2017}.

To correctly schedule a tensor operation on a hardware accelerator, one should be aware of the kind of instructions provided by the accelerator. These can be classified into two distinct categories:

\begin{itemize}
	\item \textit{Low level}: these instructions allow the programmer fine control of the behaviour of the accelerator. These are typically instructions that enable moves of data in and out of the accelerator's internal memory, and others that execute/dispatch the most basic supported computation. An example of this kind of instructions can be found on the Angel-Eye \cite{guo2018} or the VTA \cite{moreau2018} accelerators.
	\item \textit{Layer wide}: these instructions take care of the burden of managing the individual instructions, by exposing to the programmer a higher-level interface to execute an entire tensor operation, like the NVDLA \cite{nvdla}.
\end{itemize}

For a systolic array-based accelerator, we are interested in the first kind of instructions, because they allow the programmer or the tuning software fine-grained control over the generated schedules. To correctly schedule them, the code generator framework should take these 4 factors into account: 

\begin{enumerate}
	\item \textit{Configuration of the hardware}: these instructions should be generated only when the configuration of the accelerator changes, and not on each new operator, to avoid unnecessary reconfigurations.
	\item \textit{Move of data into the accelerator's SRAM}: if the input matrices fit entirely on the accelerator's internal memory, one could choose to first move the entire matrices in, and only then start to generate the compute instructions. But perhaps interleaving move and compute instructions can actually avoid idle times and thus generate faster compute schedules. The proposed schedule should be able to achieve this load balancing between different kinds of instructions.
	\item \textit{Computation}: the idle time of the systolic array should be minimized. There should always be data available in the accelerator's SRAM for the systolic array to use for its computation.
	\item \textit{Move of data out into the external DRAM}: two different options could be chosen when generating these instructions: either patches of results are moved out as soon as they are ready, or multiple patches are stored in the accumulator before bulk moving all of them out. 
\end{enumerate}

\begin{figure*}[!ht]
	\centering
	\includegraphics[width=0.81\textwidth]{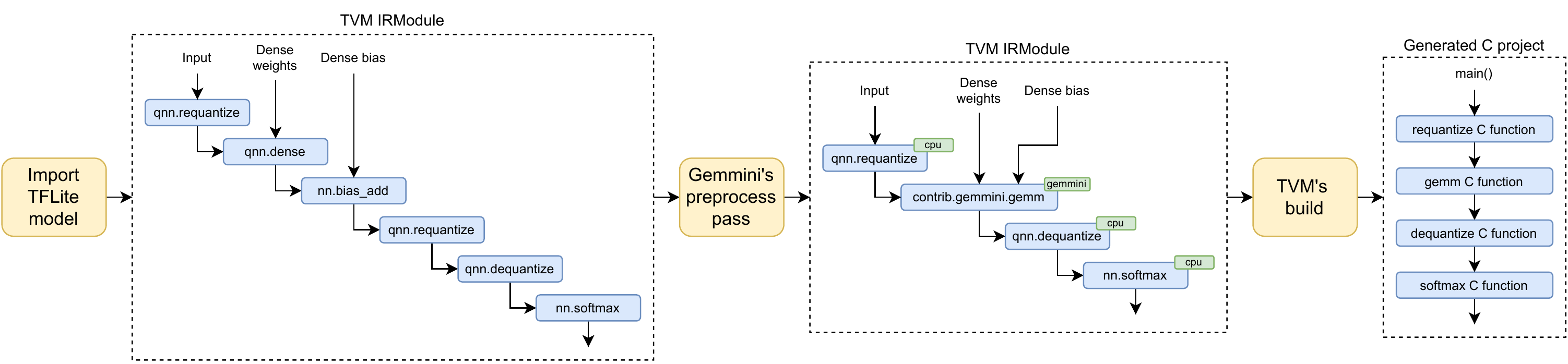}
	\caption{Integration workflow example for a neural network formed by a fully connected layer and a softmax layer}
	\label{fig:workflow}
\end{figure*}

To limit the explosion of the schedule parameter search space, certain hardware limitations and basic assumptions can be taken into account during this stage:

\begin{itemize}
	\item Schedules which do not respect the maximum limitation of columns and rows for each move should automatically be dropped.
	\item The schedule cannot generate configurations where data would overflow the accelerator's SRAM: hardware information should be used as a limit for the generated move instructions source and destination addresses.
	\item Each generated GEMM should try to utilize the systolic array to its full extent, to prevent idle processing elements.
\end{itemize}

The configuration parameters of the proposed schedule presented in Fig. \ref{fig:gemm_schedule} along with the accelerator parameters can effectively generate code that considers all the previous mentioned points:

\begin{itemize}
	\item $tile_{m,n,k}$: two-level tiling for each corresponding computation axis. The first level moves data into the accelerator's memory, and the second partitions the moved data into smaller GEMMs to fit the systolic array, trying to maximize its usage.
	\item $parallel\_accumulations$: denotes how many output patches are accumulated simultaneously in the accumulator, before moving them out. Modifies the position of the \textit{move.out} instruction in the generated code. 
	\item $apply\_double\_buffer$: represents if double buffering should be applied on the \textit{move.in} of data, weights, both or none. Takes into account the bank size of the scratchpad, so that double-buffered data is written to different banks, thus preventing bank access conflict.
	\item $exchange\_axis$: allows to reorder the computation of axis M and N, useful when M and N are not equal.
	\item $WS/OS$: if supported by the accelerator, configures the systolic array to work on output stationary mode or weight stationary mode \cite{DBLP:journals/corr/abs-1811-02883}.
	\item $mvout\_big\_block$: if supported by the accelerator, enables the generation of \textit{move.out} instructions which move more than $DIM\times DIM$ size patches. Useful to explore the trade-off between burst of smaller \textit{move.out} instructions versus bigger instructions.
\end{itemize}

\section{Integrating Gemmini into TVM}
\label{section:tvm}

To validate the aforementioned schedule search space, the TVM \cite{tvm} framework was chosen to take advantage of its AutoTVM module. This \textit{auto-tuning} process goes over the parameter search space, measures each configuration on physical hardware, and exports the best schedule configuration. To avoid an infeasible amount of measurements, XGB model-based tuners are available \cite{chen2018}.

The Gemmini \cite{gemmini-dac} systolic array-based accelerator was selected as a test platform because of its open-source nature. Gemmini was developed by the UC Berkeley and is part of the Chipyard ecosystem. It works in a tightly-coupled manner with a RISC-V CPU, using the Rocket Chip Coprocessor (RoCC) interface to control the accelerator with help from custom instructions.

Gemmini uses a systolic array of $DIM\times DIM$ \textit{multiply-and-accumulate} (MAC) processing elements to perform matrix multiplications. The data is consumed from a scratchpad made up of banked SRAMs and is stored in a series of banked SRAMs equipped with adder units known as the accumulator. A DMA engine connected to the System Bus (directly to the L2 cache) is used to get data in and out of the accelerator's SRAMs. Gemmini is also able to apply scaling factors during the move in and out of data, and also other common operations like ReLu or max pooling.

Gemmini's RoCC instructions can be grouped into the following categories:

\begin{itemize}
	\item \textit{Configuration}: these instructions configure the input, output and execution pipelines of the accelerator.
	\item \textit{Move}: these instructions move specific amounts of rows and columns of data into the SRAM or out into the DRAM.
	\item \textit{Execution}: dispatch the actual execution of the $DIM\times DIM$ GEMM to the systolic array.
	\item \textit{Flush and fence}: general maintenance instructions.
	\item \textit{Loop}: "CISC" type instructions for commonly used operations. They take away the burden of manually scheduling the intrinsic instructions by generating them directly on the hardware using FSMs. Gemmini developers also provide some handcrafted layer-wise C functions, which internally call these loop instructions.
\end{itemize}

Gemmini uses a decoupled access-execute architecture, with dedicated controllers to manage the \textit{move.in}, execute and \textit{move.out} instructions independently. A ROB is included to detect hazards between instructions and to issue them to their respective controller.

Fig. \ref{fig:workflow} shows how the developed integration between TVM and Gemmini works. First, the TensorFlow Lite quantized model is imported. Here, the model is transformed into the Relay IR dataflow graph representation of TVM. Then, pattern matching is used to replace subgraphs of operations with custom operators supported by Gemmini. These operators consist of a computation definition and its schedule: a set of parametrized loop transformations. For the GEMM operator, two different schedules were developed: one that generates the calls to the intrinsic instructions parametrized as presented in Section \ref{section:scheduling}, and one that replaces the entire loops with a call to the default handcrafted layer-wise C function.

During the schedule transformation, several \textit{pragmas} are used to tag specific loops and then replace them with the Gemmini intrinsic instructions in the following compilation pass. The \textit{tensorization} feature of TVM was used to insert the two instructions (\textit{preload} and \textit{compute}) needed by Gemmini to compute a GEMM. TVM's C code generator was used to create the source code file that executes the operators, and the standard header file provided by Gemmini was included in the generated file by TVM to correctly reference Gemmini's instruction macros.

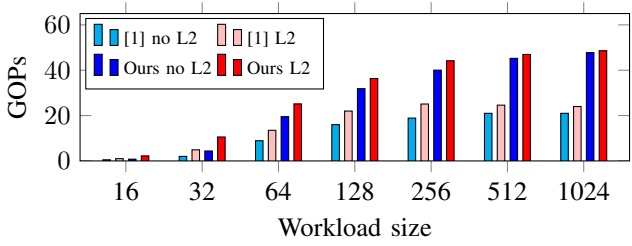
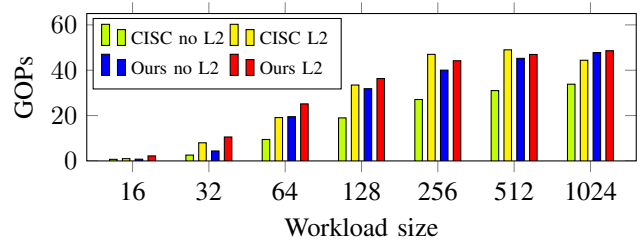
\begin{figure*}[!ht]
	\centering
	\begin{subfigure}[b]{0.49\textwidth}
		\begin{tikzpicture}
			\begin{axis}[
				xlabel=Workload size,
				ylabel=GOPs,y unit= ,
				width=\textwidth,
				height=0.15\textheight,
				bar width=0.1cm,
				ybar,
				xtick=data,
				ymin=0,
				ymax=65,
				legend style={legend columns=2},
				legend style={at={(0.444,0.97)}},
				legend style={font=\scriptsize},
				legend cell align={left},
				symbolic x coords={
					16,
					32,
					64,
					128,
					256,
					512,
					1024}]
				\addplot[fill=cyan] table[x=workload,y=gops,col sep=comma] {./data/xu2020_autotvm_nocache.csv};
				\addplot[fill=pink] table[x=workload,y=gops,col sep=comma] {./data/xu2020_autotvm_cache.csv};
				\addplot[fill=blue] table[x=workload,y=gops,col sep=comma] {./data/ours_autotvm_nocache.csv};
				\addplot[fill=red] table[x=workload,y=gops,col sep=comma] {./data/ours_autotvm_cache.csv};
				\addlegendentry{[1] no L2}
				\addlegendentry{[1] L2}
				\addlegendentry{Ours no L2}
				\addlegendentry{Ours L2}
			\end{axis}
		\end{tikzpicture}
		\caption{Best schedules found by AutoTVM in our implementation against previous work.}
		\label{fig:xu2020_vs_ours}
	\end{subfigure}
	\hfill
	\begin{subfigure}[b]{0.49\textwidth}
		\begin{tikzpicture}
			\begin{axis}[
				xlabel=Workload size,
				ylabel=GOPs,y unit= ,
				width=\textwidth,
				height=0.15\textheight,
				bar width=0.1cm,
				ybar,
				xtick=data,
				ymin=0,
				ymax=65,
				legend style={legend columns=2},
				legend style={at={(0.464,0.97)}},
				legend style={font=\scriptsize},
				legend cell align={left},
				symbolic x coords={
					16,
					32,
					64,
					128,
					256,
					512,
					1024}]
				\addplot[fill=lime] table[x=workload,y=gops,col sep=comma] {./data/ours_baseline_nocache.csv};
				\addplot[fill=yellow] table[x=workload,y=gops,col sep=comma] {./data/ours_baseline_cache.csv};
				\addplot[fill=blue] table[x=workload,y=gops,col sep=comma] {./data/ours_autotvm_nocache.csv};
				\addplot[fill=red] table[x=workload,y=gops,col sep=comma] {./data/ours_autotvm_cache.csv};
				\addlegendentry{CISC no L2}
				\addlegendentry{CISC L2}
				\addlegendentry{Ours no L2}
				\addlegendentry{Ours L2}
			\end{axis}
		\end{tikzpicture}
		\caption{Best schedules found by AutoTVM in our implementation against the default "CISC instruction" based schedules.}
		\label{fig:ours_vs_cisc}
	\end{subfigure}
	\caption{Results across different GEMM workloads. For each workload, $M=N=K=workload\:size$}
	\label{fig:result}
\end{figure*}

\subsection{Quantization management}

TensorFlow Lite's quantization scheme \cite{DBLP:journals/corr/abs-1712-05877} uses symmetric quantization for the weights and biases of each layer, and asymmetric for the input and output of each layer \footnote{see \url{https://www.tensorflow.org/lite/performance/quantization_spec?hl=en}}. Because the multiplication of two quantized matrices with different zero points is not straightforward, correction terms have to be subtracted to get the correct result as seen in Eq. \eqref{eq:qgemm_terms}. Finally, to transform $Q'_{C}$ into the output quantization regime of the layer, a \textit{requantization} operator is inserted by TVM described by Eq. \eqref{eq:req}.

An easy approach would be to accelerate only the matrix multiplication using the systolic array (the term 1 of Eq. \eqref{eq:qgemm_terms}) and then execute the correction terms subtraction and the re-quantization operator on the CPU. But this is slow and doesn't exploit the benefits of TVM's compilation framework, so a transformation of the GEMM operation is done during the compilation pass, allowing TVM to fold the remaining constants into the bias of the layer. The final proposed solution for a quantized matrix multiplication with bias addition is described in Eq. \eqref{eq:qgemm_corrected} and \eqref{eq:new_bias}. The scaling factor $s_d/s_c$ is inserted as output scaling factor in the configuration of the \textit{move.out} instruction of Gemmini, thus achieving the acceleration of the entire operator, without extra tensor operations executed on the CPU.

{\scriptsize
	\begin{multline}
		\label{eq:qgemm_terms}
		Q'_{C_{(m,n)}} = \sum_{0}^{k}\left(Q_{A_{(m,k)}}-zp_a\right) * Q_{B_{(k,n)}} = \\ \sum_{0}^{k}Q_{A_{(m,k)}}*Q_{B_{(k,n)}} - \sum_{0}^{k}zp_a*Q_{B_{(k,n)}}
	\end{multline}
	
	\begin{align}
		\label{eq:req}
		Q_{C} = zp_{c} + \frac{s_{d}}{s_{c}} * Q'_{C} \\
		\label{eq:qgemm_corrected}
		Q_{C_{(m,n)}} = \frac{s_{d}}{s_{c}}\left[\left(\sum_{0}^{k}Q_{A_{(m,k)}}*Q_{B_{(k,n)}}\right)+Q'_{D_{(m,n)}}\right]\\
		\label{eq:new_bias}
		Q'_{D_{(m,n)}} = Q_{D_{(m,n)}}-\sum_{0}^{k}zp_a*Q_{B_{(k,n)}}+\frac{s_{d}}{s_{c}}*zp_{c}
	\end{align}
}

\section{Experiments}
\label{section:experiments}

{\scriptsize
	\begin{table}[htbp]
		\caption{Selected Baidu DeepBench workloads}
		\begin{center}
		\begin{tabular}{cccc} \toprule
			Id & M & N & K \\
			\midrule
			15 & 64 & 1 & 1216\\
			49 & 128 & 1 & 1024\\
			63 & 512 & 1 & 512\\
			73 & 512 & 2 & 512\\
			84 & 1024 & 4 & 512\\
			\bottomrule
		\end{tabular}
		\label{tab:baidu_workloads}
		\end{center}
	\end{table}
}

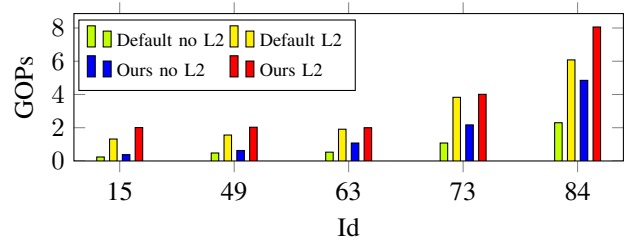
\begin{figure}[!ht]
	\centering
	\begin{tikzpicture}
		\begin{axis}[
			xlabel=Id,
			ylabel=GOPs,y unit= ,
			width=0.49\textwidth,
			height=0.15\textheight,
			bar width=0.1cm,
			ybar,
			xtick=data,
			ymin=0,
			legend style={legend columns=2},
			legend style={at={(0.513,0.96)}},
			legend style={font=\scriptsize},
			legend cell align={left},
			symbolic x coords={
				15,
				49,
				63,
				73,
				84}]
			\addplot[fill=lime] table[x=workload,y=gops,col sep=comma] {./data/ours_baidu_baseline_nocache.csv};
			\addplot[fill=yellow] table[x=workload,y=gops,col sep=comma] {./data/ours_baidu_baseline_cache.csv};
			\addplot[fill=blue] table[x=workload,y=gops,col sep=comma] {./data/ours_baidu_autotvm_nocache.csv};
			\addplot[fill=red] table[x=workload,y=gops,col sep=comma] {./data/ours_baidu_autotvm_cache.csv};
			\addlegendentry{Default no L2}
			\addlegendentry{Default L2}
			\addlegendentry{Ours no L2}
			\addlegendentry{Ours L2}
		\end{axis}
	\end{tikzpicture}
	\caption{Best schedules found by AutoTVM for the Baidu DeepBench dataset using our implementation.}
	\label{fig:ours_vs_cisc_baidu}
\end{figure}

In order to be able to compare our results against the ones reported by \cite{xu2020}, our test setup was also built using the Rocket Core \cite{Asanovic:EECS-2016-17} together with a $16\times16$ Gemmini accelerator, with a 256 KB scratchpad (4 SRAM banks with 4096 rows each) and a 64 KB accumulator (1 SRAM bank with 1024 rows). The inputs and weights are represented using 8 bits and the accumulated values using 32 bits. In the implemented design which uses the L2 cache, the SiFive inclusive L2 cache was used. Like the original paper, the resulting hardware was also implemented for a Xilinx Zynq UltraScale+ ZCU102 FPGA running at 100 MHz. The only difference between our setup and the one from \cite{xu2020} is the size of the L2 cache because they did not report the selected size. In our work, the default size used in the Chipyard project was selected: 512 KB.

Fig. \ref{fig:result} presents the best schedule performance found by AutoTVM, and compares it with the previous work and the default Gemmini schedules. An XGB tuner \cite{chen2018} with early stopping equal to 500 iterations was used to traverse the search space. The amount of operations for a generic GEMM of form $C_{[M,N]} = A_{[M,K]}\times B_{[K,N]}+D_{[M,N]}$ was defined as $OP = 2\times M\times N \times K + M\times N$.

Although Fig. \ref{fig:ours_vs_cisc} shows that our AutoTVM schedules are better than the CISC-based schedules for almost all workloads, it fails to find better schedules for workloads 256 and 512 when the L2 cache is activated. There are two possible explanations for this behaviour. The first one would be that the XGB tuner is stuck in a local optimum, but this was verified to \textbf{not} be the case by executing an exhaustive grid-search of all the schedule parameter search space: the XGB found schedules are indeed the best possible schedules our implementation can generate for that workloads. The second possible explanation lies in a \textit{load balancing} feature of the Gemmini's CISC instructions. The hardware FSMs monitor the proportion of each instruction type in the ROB and can pause the generation of each kind of instruction, to maximize the overlap between move and execute operations. This behaviour can not yet be replicated using our TVM integration, and further work needs to be done to analyse how to implement a similar feature.

To show the effectiveness of this schedule search space on real-world examples, a set of dense layer workflows taken from the Baidu DeepBench dataset \cite{narang2017baidu} were selected (Table \ref{tab:baidu_workloads}). Fig. \ref{fig:ours_vs_cisc_baidu} presents the result of the AutoTVM tuning process executed on these workloads, using the same tuner parameters and FPGA bitstreams as in the previous measurements.

\section{Conclusions}
\label{section:conclusions}

This work proposed a schedule parameter space for a GEMM tensor operation. The proposed schedule was configured into the TVM deep learning compiler, and integrated with the Gemmini systolic array hardware accelerator. We demonstrate that this schedule parameter space allows the autotuning process to find faster schedules than previous work, and also faster schedules than the hardcoded schedules provided by the expert developers of Gemmini on almost all tested workloads.

Future works will also add other operators to the integration, like convolutions and depthwise convolutions. The results of the acceleration of entire neural networks will also be presented.

This work did not try to improve how much time the autotuning process takes. The search space of scheduling parameters was traversed using an XGB tuner approach, which can be improved to converge faster as shown in \cite{reiter2022}. In future works, optimization techniques to speed up this process should be investigated, to be able to find the best scheduling parameters using the minimum amount of actual hardware measurements.

\balance

\bibliographystyle{IEEEtran}
\bibliography{bib}

\end{document}